\documentclass{svproc}   
\usepackage{footmisc} 										
\usepackage{multicol} 										

\usepackage{subfigure}

\newcommand{\bigO}{\mathcal{O}}
\newcommand{\road}{\mathcal{R}}

\newcommand{\OPEN}{\textsc{Open}}
\newcommand{\CLOSED}{\textsc{Closed}}

\newcommand{\node}{n}
\newcommand{\nodeI}{n_{\mathrm{I}}}

\usepackage[T1]{fontenc}  								        
\usepackage[applemac]{inputenc}							         
\makeatletter
\let\l@English\l@english
\makeatother
\usepackage[english]{babel} 								        
\usepackage[babel]{csquotes}									
\usepackage{lipsum} 										
\usepackage{url} 											
\usepackage{fancyhdr}										

\usepackage{emptypage}										
\usepackage[english]{varioref} 							                  
\usepackage{microtype}										
\usepackage{eurosym}										
\usepackage{lettrine}                                                                                            
						
\usepackage[fleqn]{amsmath}

\usepackage{mathrsfs}										
\usepackage{amsfonts}										
\usepackage{amssymb}

\usepackage{graphicx}
\usepackage[linesnumbered,ruled,vlined]{algorithm2e}
\usepackage{epstopdf}
\usepackage{upgreek}	
\usepackage{paralist}

\usepackage{mathtools}

\newenvironment{sistema}
{\left\lbrace\begin{array}{@{}l@{}}}								 
{\end{array}\right.}
\usepackage{amscd}										 
\usepackage{booktabs}										 
\usepackage{graphics} 										 
\usepackage{epsfig} 										 
\usepackage{mathptmx} 										 
\usepackage{times} 											 
\usepackage{rotating}										 
\usepackage{tabularx}										
\usepackage{longtable}										 
\usepackage{pdflscape}										 
\usepackage{listings}										 
\usepackage{xcolor} 										 
\usepackage{siunitx}

\DeclareMathAlphabet{\mathpzc}{OT1}{pzc}{m}{it}
\DeclareMathAlphabet{\mathcal}{OMS}{cmsy}{m}{n}

\usepackage{cite}
\usepackage{balance}
\usepackage{tikz}				% figure
\usetikzlibrary{arrows,shapes,backgrounds,calc,positioning,patterns}
\usepackage{multirow}
\usepackage{cases}
\usepackage{fancyhdr}

\fancypagestyle{plain}{}

\usepackage[moderate]{savetrees}
\begin{document}

\title{Search-Based Motion Planning \\ for Performance Autonomous Driving (DRAFT)} %\thanks{Work supported by EU Project ITEAM (project reference: 675999).}}
\titlerunning{Search-Based, Performance Oriented Motion Planning (DRAFT)}  % abbreviated title (for running head)

\author{Zlatan Ajanovic\inst{1}\thanks{Equal contributions. Ajanovic focused on algorithmic while Regolin on vehicle dynamics part.}, Enrico Regolin\inst{2}$^{\ast}$, \\ Georg Stettinger\inst{1}, Martin Horn\inst{3} and Antonella Ferrara\inst{2}}
\institute{Virtual Vehicle Research Center, %Inffeldgasse 21/A, 8010 Graz, 
Austria, \\
\email{\{zlatan.ajanovic, georg.stettinger\}@v2c2.at},
\and
Dipartimento di Ingegneria Industriale e dell'Informazione, University of Pavia, %via Ferrata 5, 27100 Pavia,
Italy, \\
\email{enrico.regolin@unipv.it, a.ferrara@unipv.it}
\and
Graz University of Technology, %Inffeldgasse 21/B, 8010 Graz, 
Austria, 
\email{martin.horn@tugraz.at}}
\authorrunning{Ajanovic, Regolin et al.} % abbreviated author list (for running head)

\maketitle

{}
%************************************************************* ABSTRACT *******************************************%
\begin{abstract}
Driving on the limits of vehicle dynamics requires predictive planning of future vehicle states. In this work, a search-based motion planning is used to generate suitable reference trajectories of dynamic vehicle states with the goal to achieve the minimum lap time on slippery roads. The search-based approach enables to explicitly consider a nonlinear vehicle dynamics model as well as constraints on states and inputs so that even challenging scenarios can be achieved in a safe and optimal way. The algorithm performance is evaluated in simulated driving on a track with segments of different curvatures. 

\end{abstract}

%%%%%%%%%% (I) INTRODUCTION
\section{Introduction}
\label{sec:intro}
\noindent
Similarly to other autonomous systems, autonomous vehicle control is based on a Sensing-Planning-Acting cycle. 
Predictive planning of future vehicle states can enable real-time control of driving while avoiding static obstacles \cite{liniger2015optimization} and can be extended to racing scenarios with multiple agents (although not real-time) \cite{liniger2019noncooperative}. 
However, this Motion Planning (MP) approach is based on an exhaustive search and works well only for short horizons (due to exponential complexity) and in high friction conditions (where fast transitions between simple, constant velocity primitives can be achieved). Approaches like this cannot be used for controlling a vehicle in lower friction conditions, which require larger horizons and a more detailed vehicle model as 
the control action often enters the saturated region.

Another line of work considers this kind of situations, i.e. driving with high side-slip angles like drifting, trail-braking, etc.
Most of the current work in this direction considers sustained drift or transient drift scenarios.
% Zico Kolter
One example of a transient drift scenario is drift parking, as shown in \cite{kolter2010probabilistic}, where the vehicle enters temporarily a drift state.
%drifting 
On the other hand, in sustained drift scenarios, the goal is to maintain steady-state drifting. Velenis et al. modeled high side-slip angle driving and showed that for certain boundary conditions it can be a solution for the minimum-time cornering problem \cite{velenis2015modeling}, \cite{velenis2008optimality}. Tavernini et al. showed that aggressive drifting maneuvers provide minimum time cornering in low-friction conditions \cite{tavernini2013minimum}.
Because of computational complexity, most of these works cannot achieve online performance.
% Major

Based on results generated offline, by using \cite{velenis2008optimality}, You and Tsiotras proposed a solution for the learning of primitive trail-braking behavior, enabling online generation of trail-brake maneuvers \cite{you2018real}. This approach decomposes trail-braking into three stages: entry corner guiding, steady-state sliding and straight line exiting.
%Borrelli
A similar decomposition of the problem was also presented in \cite{zhang2018drift}. This one divides the horizon into three regions, finds a path for each region 
%(using Rapidly-exploring Random Trees (RRT), rule-based sampling and Proportional Integral Control) 
and then concatenates them. As it appears from the results, though,  
in the drifting region the rule-based solution produces non optimal solutions.
%conslusion on these 2
Impressive demonstration of model-based reinforcement learning approach on scaled vehicle is shown in \cite{7989202}, however, as for the aforementioned sustained drift approaches, considered scenario is relatively simple with only one curve. It is hard to expect that these approaches generalize well to more complicated scenarios. Driving the full track generally requires solving more curves, with variable curvature radius and a mix of right and left curves. 
In this scenario the \lq\lq three-segments'' assumption does not hold, and finding out how to split the horizon and assigning the segments arises as a problem, which can be viewed as a combinatorial optimization problem. This confirms that the problem of continuous driving is a different one from sustained drift or transient drift. 
%driving full track 

% Our method
As generating references in a continuous driving problem can be considered as a combinatorial optimization problem, heuristic search methods like $A^{\ast}$ can be effectively used for automated optimal trajectory generation.
In this paper, a novel $A^{\ast}$ search-based approach to the problem of generating \lq\lq trackable'' references is presented.
The presented $A^{\ast}$ search-based planner is a modified version of the one illustrated in \cite{ajanovic2018search}, where a rather simple vehicle model was used to generate vehicle trajectories in complex urban driving scenarios. % (also combinatorial problem). 
The space of the possible trajectories is explored by expanding different combinations of motion primitives in a systematic way, guided by a heuristic function.
Motion primitives are generated using two different vehicle models. A bicycle model is used for small side-slip angle operations (i.e. entry and exit maneuvers and close-to-straight driving) and a full nonlinear vehicle model for steady state cornering maneuvers.
Such automated motion primitives generation enables to generate arbitrary trajectories, not limited to just one curve as in previous approaches.

%closed loop prediuction
The presented expansion approach is similar to the closed-loop prediction approach (CL-RRT) \cite{kuwata2008motion}, which uses closed-loop motion primitives for the expansion when open-loop dynamics are unstable, so that the exploration by variations in the open-loop dynamics becomes inefficient. In our approach, we expand steady state cornering motion primitives, which are considered to be achievable via closed-loop control such, as in e.g. \cite{regolin2018multi}.

% %OLD

%%%%%%%%%% (II) Vehicle Models
\section{Problem Formulation and Vehicle Models}
\label{sec:problem_models}

The tackled problem in this work is minimum lap time driving on an empty track in low friction conditions, i.e. gravel road. % in the considered usecase. 
%To formulate this problem it is necessary to define a vehicle model and the race track. 
It is assumed that the vehicle is equipped with a map of the road and a localization system. Therefore, the vehicle has the information about the road ahead, as well as left/right boundaries and exact position and orientation.
Moreover, the full vehicle state feedback information is available. In particular, besides dynamic states, the low level controller (which is not covered in this work) for state tracking requires measurements and estimates of several quantities, including wheel forces and wheel slips, both longitudinal and lateral (see e.g. \cite{regolin2019sliding}).  
Finally, the combined longitudinal/lateral tire-road contact forces characteristics are assumed to be known and constant. 
%road
The road, on the other hand, is assumed to be flat, with static road-tire characteristic and can have arbitrary shape with constant width.

\subsection{Vehicle kinematic model}
The vehicle trajectories are generated by concatenating smaller segments of trajectories, the so-called "motion primitives". These are generated based on a model for the vehicle planar motion.
Such model is comprised of six states
$[x, y, \psi, v, \beta, \dot\psi]^{T}$, where $x$, $y$ and $\psi$ represent kinematic states (position and yaw angle), and $v$, $\beta$ are vehicle velocity and side-slip angle respectively.
The evolution of $x$ and $y$ is given by:
\begin{equation}
\begin{sistema}
 \dot{x} = v \cdot cos(\psi+\beta)\\
 \dot{y} = v \cdot sin(\psi+\beta)
\label{eq:kinematic_model}
\end{sistema}
\end{equation} 
whereas $ v$, $\beta$, $\dot\psi$ are given by the selected vehicle model.

In regular driving situations, i.e. for small values of $\beta$, the trajectory identified by \eqref{eq:kinematic_model} is mostly determined by $v$ and $\dot\psi$, and therefore can be planned by means of linearized vehicle models, valid for small variations of  $\beta$ and $\dot\psi$. 
%In \cite{ajanovic2018search} the space of the possible trajectories is explored by expanding different combinations of longitudinal and lateral motions.
When driving on slippery surfaces, such solution is not suitable anymore, due to the effect of $\beta$ in \eqref{eq:kinematic_model} and to the complexity of the model which describes the vehicle motion for growing values of $\beta$. 
A major limitation which stems from using a full nonlinear vehicle model is the so-called "curse of dimensionality", and related computational explosion when introducing new states.
In fact, if motion primitives were generated with the full nonlinear vehicle model, the computational burden would increase excessively, thus making it a non viable option for an online implementation.
%When modelling vehicle dynamics, the so-called linearized \lq\lq bicycle model'' is one of the most widely used ones. However, when the limits of handling are reached, its approximations do not allow it to properly account for the nonlinear phenomena occurring at wheel-road contact level.

To overcome unnecessary increase in computation, we generate motion primitives by using: 
\begin{inparaenum}[i)]
\item the so-called \lq\lq bicycle model'' \cite{genta1997motor} in straight-driving/mild-turning scenarios, and 
\item a convenient approximation of the full nonlinear model, based on the theoretical vehicle-equilibrium-states during cornering.
\end{inparaenum}

\subsection{Equilibrium States Manifold}
\label{subsec:v_manifold}
%zlatan added sentence
For cornering, the motion primitives are generated based on previously computed vehicle-equilibrium-states. % from equilibrium states manifold.
When generating target reference states, one of the main issues is ensuring that the reference values can be actually reached in a sufficiently short time. This problem is made harder by the slow dynamics of the vehicle longitudinal and lateral accelerations, due to the slippery surface considered. Two devices are exploited to counter this problem: 
\begin{inparaenum}[i)]
\item \lq\lq slow varying'' reference set-points are used, in order to allow the low level actuator to bring the actual state in proximity of the desired one; 
\item only reference states belonging to a specific \lq\lq Equilibrium States Manifold'' (ESM) are considered. 
\end{inparaenum}
To obtain such manifold, which requires a reliable model for vehicle and tire-road contact forces, an off-line computation is performed, which provides the steady-state solutions of the vehicle cornering at different curvature radii \cite{velenis2011steady}. These solutions include the vehicle control inputs (steering wheel angle and rear wheels slip) as well as the vehicle states $v$, $\beta$, $\dot\psi$.
For the offline computation of the equilibrium points, the full-vehicle nonlinear model is exploited, where the tire-road forces $F_{x},F_{y}$ for each axle ($F_{x,f}=0$ due to the RWD configuration) are obtained from the normal forces $F_z$ and the combined longitudinal/lateral friction model  (see \cite{pacejka2012}):
\begin{align}
\label{eq:forces_definition}
F_{x,i}=F_{z,i} \mu_x(\lambda_i,\alpha_i),\quad F_{y,i}=F_{z,i}\mu_y(\lambda_i,\alpha_i)
\end{align}
where $\mu$ is the friction coefficient, $\lambda$ and $\alpha$ the longitudinal and lateral slips respectively.

The nonlinear friction functions take the form of the \textit{Magic Formula} (MF) tire friction model, with an isotropic friction model being used for simplicity. This requires the computation of the theoretical slip quantities ($\sigma_j$, $j\in\{x,y\}$), which can be obtained from $\alpha$, $\lambda$ as follows:
\begin{equation}\label{eq:2_8}
\sigma_{x}=\frac{\lambda}{1+\lambda},\quad \sigma_{y}=\frac{\tan \alpha}{1+\lambda},\quad \sigma=\sqrt{\sigma_x^2+\sigma_y^2}
\end{equation}   
Then, the  one-directional friction coefficients are given by
\begin{align}
\begin{split}
\mu_i&=\frac{\sigma_i}{\sigma}D  \sin[C_\lambda \arctan \{\sigma B  - E  (\sigma B-\arctan \sigma B )\}] \mathrm{,}
\label{eq:mu_def}
\end{split}
\end{align}
for $i=x,y$, with $B=1.5289$, $C=1.0901$, $D=0.6$, $E=-0.95084$ being the Pacejka parameters corresponding to gravel. 

\begin{figure}[!tp] 
\centering
\subfigure{
%\subfigure[ESM - $\delta$]{%
\includegraphics[width=.47\columnwidth]{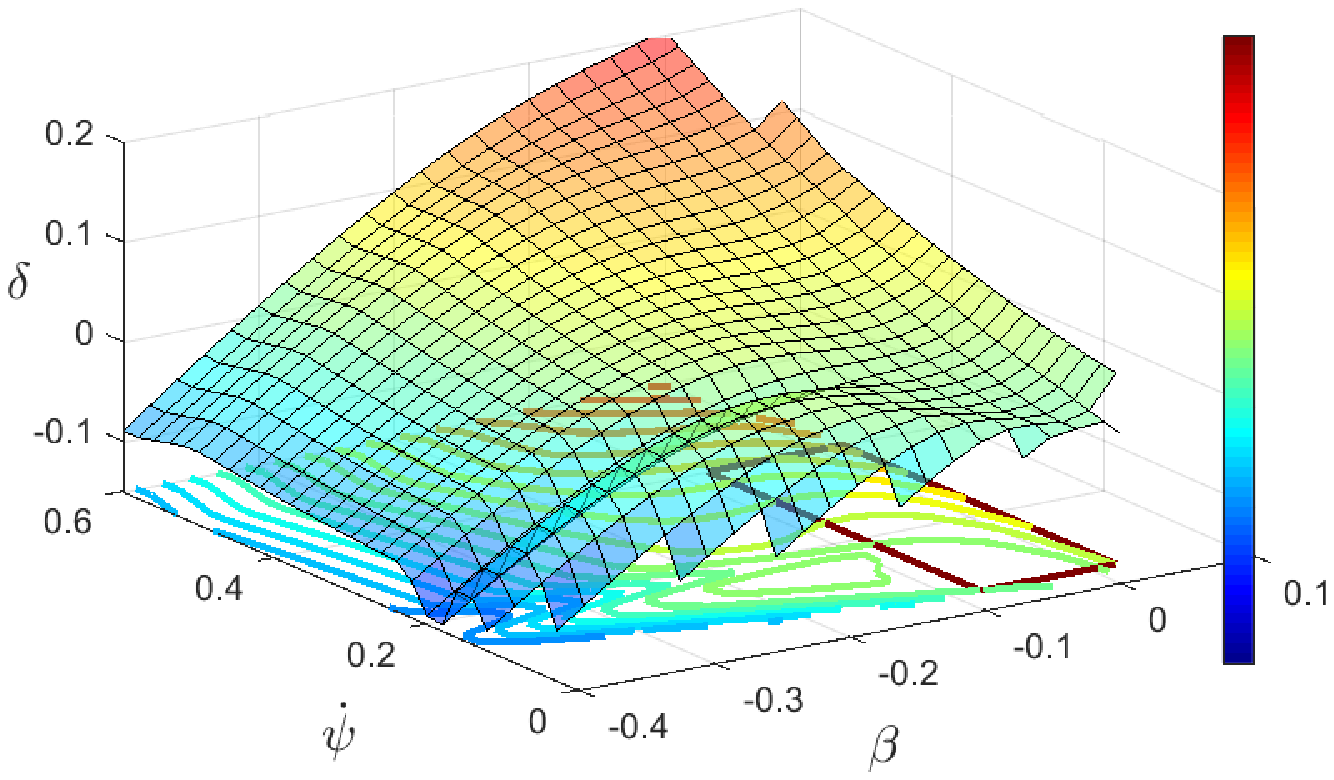}%
%\label{fig:surface_delta}%
}\hfil
\subfigure{
%\subfigure[ESM - $\lambda$]{%
\includegraphics[width=.47\columnwidth]{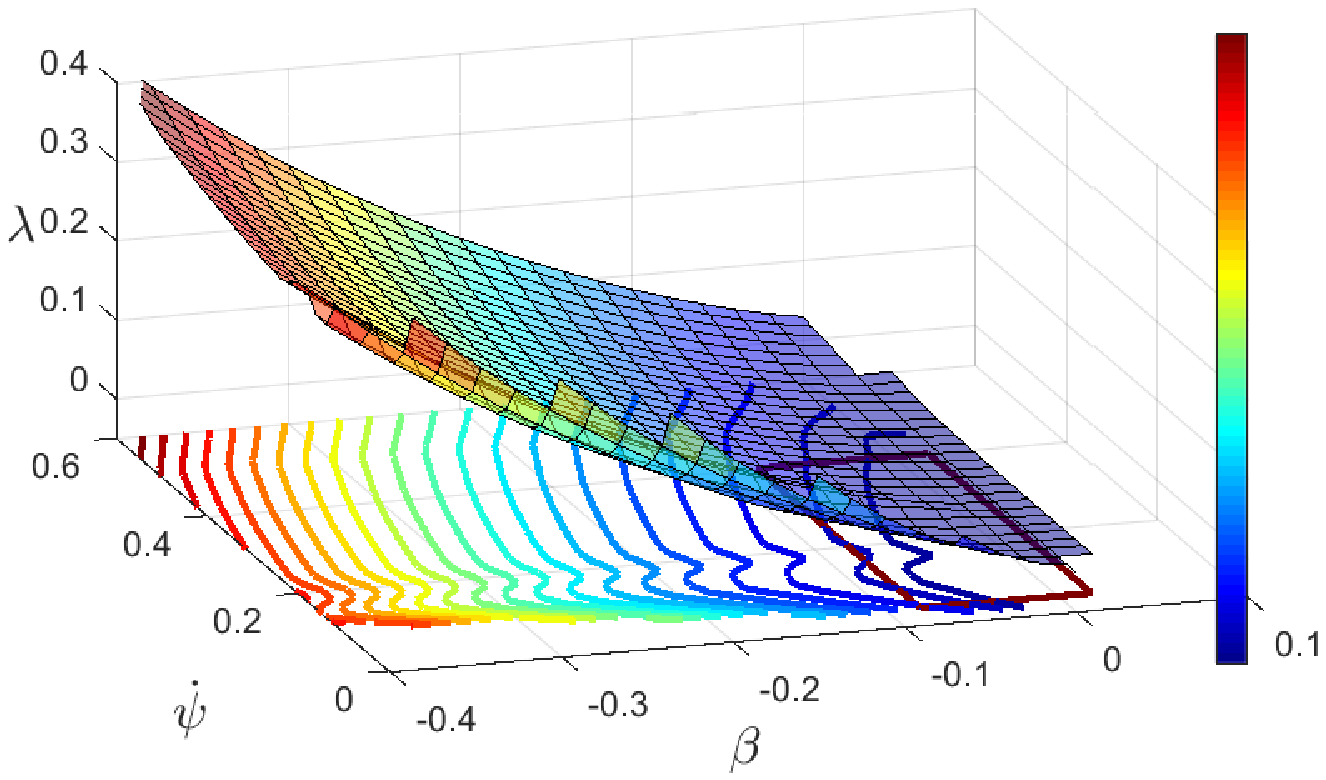}
%\label{fig:surface_lambda}%
}
\caption{ESM for the inputs $\delta$ (left) and $\lambda$ (right), counter-clockwise maneuvers. The highlighted portion of the $\beta-\dot\psi$ plane corresponds to the one in which the bicycle model representation is considered valid. The intervals of values $[\delta_{min},\delta_{max}]$, $[\lambda_{min},\lambda_{max}]$ are used for the bicycle-model expansion explained in Section \ref{subsec:node_exp}.}
\label{fig:surfaces_inputs}%
\end{figure}

The vehicle responses %(vehicle velocity $v$, side-slip angle $\beta$ and yaw rate $\dot\psi$) 
can be obtained from the following system of nonlinear equations, which considers the lateral, longitudinal and rotating balance equilibrium equations around the vehicle center of gravity:
\begin{equation}
\begin{sistema}
 \dot{v} = \dot{\psi}v\beta+\frac{F_{x,r}}{m}     \\
m v (\dot{\beta}+\dot{\psi}) = F_{y,f}+F_{y,r}\\%-C_f (\beta + \frac{l_}{v}\dot{\psi}-\delta)-C_r (\beta - \frac{l_r}{v}\dot{\psi})  \\
J_z \ddot{\psi} = l_fF_{y,f}-l_rF_{y,r}%-l_fC_f (\beta + \frac{l_}{v}\dot{\psi}-\delta) +l_rC_r (\beta - \frac{l_r}{v}\dot{\psi}) + M_{\dot{\psi}}  
\label{eq:nonlinear_model}
\end{sistema}
\end{equation} 
where $J_z$ is the vehicle inertia around the z-axis, $m$ the vehicle mass and $l_f$, $l_r$ are the distances of the vehicle COG from the front and rear axles respectively. %Besides \eqref{eq:nonlinear_model}
In addition, also the longitudinal weight transfer is considered.
\begin{figure}[!tp]
\begin{center}
\includegraphics[width=.8\columnwidth]{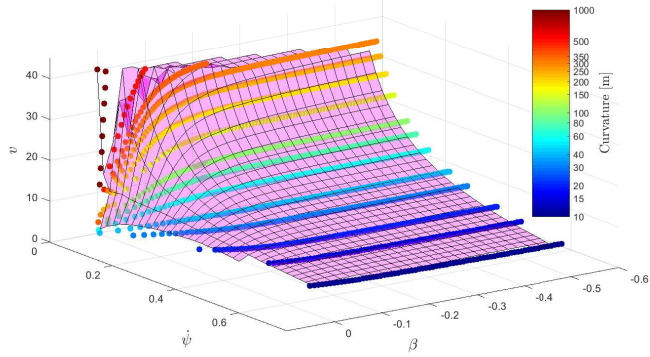}
\caption{Equilibrium points sets $S_{ss}$ (and linear interpolation) in the $v,\beta,\dot\psi$ space for counter-clockwise cornering maneuvers with different curvature radii $R_c$. }
\label{fig:surface_interp}
\setlength{\belowcaptionskip}{-10pt}  
\setlength{\abovecaptionskip}{-10pt}
\end{center}
\end{figure}
\noindent
Assuming a RWD drivetrain configuration, and given different sets of values of the constant control inputs (steering wheel angle $\delta$, driving wheels slip $\lambda$)  the equilibrium points $S_{ss} = [v_{ss}, \beta_{ss}, \dot\psi_{ss}]^{T}$ can be computed, in order to achieve different constant curvature radii $R_c$, by considering the uniform circular-motion relation $\dot\psi = \frac{v}{R_c}$, 
and imposing in \eqref{eq:nonlinear_model} the steady-state condition 
\begin{equation}
\label{eq:derivatives_0}
\dot{v}=\dot\beta= \ddot\psi = 0\mathrm{.}
\end{equation} 
In Fig. \ref{fig:surface_interp} these sets are graphically displayed in the 3-dimensional state-space, for varying $R_c$.
The corresponding surfaces, generated for $\delta$ and $\lambda$ are displayed in Fig. \ref{fig:surfaces_inputs}.

%======================================%
%======================================%

Let us assume that the tire-road contact model and the vehicle model \eqref{eq:forces_definition}-\eqref{eq:nonlinear_model} 
% describe correctly the cornering maneuver dynamics
are correct, and that a trajectory of any curvature radius $R_c$ is given, for which at least one reference state  $S_{ss}(R_c)$ exists. Then, if a locally stable feedback controller for the tracking of the state is designed, such trajectory can be tracked, given an initial condition close enough to the target state.

%One can notice how, although this is a sufficient condition for path tracking in case of steady-state cornering, a
A race track is composed of different sections, with varying  curvature radii as well as straight segments. Therefore, in order to determine the sequence of vehicle states to be tracked by the vehicle, a continuous transition between different equilibrium points would be required, neglecting condition \eqref{eq:derivatives_0}.
For this reason, the sets $S_{ss}(R_{c_i})$, $i=1,..,r$ in Fig. \ref{fig:surface_interp} have been interpolated into a map $v=f(\beta,\dot{\psi})$, which represents the ESM from which the sequences $v,\beta,\dot\psi$ which generate the trajectories are taken, the same procedure being applied for $\delta$, $\lambda$ in Fig. \ref{fig:surfaces_inputs}. 

\subsection{Drivable road and Criteria}

It is considered that the vehicle can drive only on the road $\road$. %The states outside of the road are considered non-drivable states and they are treated as obstacles $(\bigO)$ by the motion planning algorithm. 
Therefore, the drivability of the trajectory generated based on the vehicle model is validated based on vehicle coordinates $x$, $y$ and yaw angle $\psi$ only (not by higher dynamic states).
To simplify planning, the drivable road is modeled using a Frenet frame \cite{werling2012optimal}. Instead of using $x$ and $y$ coordinates, in Frenet frame one dimension represents the distance traveled along the road $s$, and other the deviation from the road centerline $d$. In this way, to determine whether the vehicle is on the road, it is sufficient to check the lateral deviation in Frenet frame. The Frenet frame is only used for trajectory evaluation during planning, %  and grid in planning such that 
to ensure that the 
planning procedure remains the same for each segment of the road. It is not used for the underlying motion primitive generation, which is based on the vehicle dynamic model in Euclidean space.

As for the trajectory evaluation criteria,
since the goal is to minimize lap time and planning horizon time is fixed, an equivalent behavior can be achieved by maximizing the distance traveled along the road in a defined time horizon. The criteria can be evaluated simply by considering the first coordinate in Frenet frame. 

%%%%%%%%%% (III) A star Algorithm Description
\section{Motion planning}

\begin{algorithm}[t]
\DontPrintSemicolon
\fontsize{8pt}{9pt}\selectfont
	\SetKwData{n}{$n$}
	\SetKwFunction{Expand}{Expand}
	\SetKwFunction{Col}{ColCheck}
	\SetKwFunction{Select}{Select}
	\SetKwFunction{Children}{Children}
	\SetKwFunction{Path}{Path}
	\SetKwInOut{Input}{input}\SetKwInOut{Output}{output}
	\Input{$\nodeI$, Obstacles data $(\bigO)$, $h(\node)$} %, \Expand(n), \Col(n)}
	\BlankLine
	\Begin{
 		$\node \gets \nodeI, \CLOSED \gets \varnothing, \OPEN \gets \node$\tcp*[r]{initialization}
		\BlankLine		
		\While{$n.k \leq k_{hor} $ {\bf and} $ \OPEN \neq \varnothing $ {\bf and}  $ \OPEN.size() \leq N_\mathrm{timeout} $}{
			%\BlankLine
			$n \gets \Select(\OPEN)$\;
			%\BlankLine
			$( n', n_{\mathrm{C}})\gets $  \Expand{$n$, \Children, \Col}\; 			
			%\BlankLine
			$\CLOSED \gets \CLOSED \cup n \cup n_{\mathrm{C}}, \OPEN \gets \OPEN \setminus n \cup n'$\;			
		}
		\Return{\Path()}\;
	}
\caption{$A^{\ast}$ Heuristic Search for a horizon\label{alg1}}
\end{algorithm}
The proposed Motion planning framework is based on the $A^{\ast}$ search method \cite{hart1968formal}, guided by an heuristic function in an MPC-like replanning scheme. After each time interval $T_{rep}$, replanning is triggered from the current vehicle state, together with information about the drivable road ahead. 

The trajectory is generated by a grid-like search using $A^{\ast}$ (see Algorithm \ref{alg1}). The grid is constructed via discretization of the state variables $\textbf{x}$. Starting from the \textit{initial node} (i.e. state), chosen as the first \textit{current node}, all neighbors are determined by expanding the \textit{current node} using motion primitives. The resulting \textit{child nodes} are added to the $\OPEN$ list. If the \textit{child node} is already in the $\OPEN$ list, and the new \textit{child node} has a lower cost, the parent of that node is updated, otherwise it is ignored. From the $\OPEN$ list, the node with the lowest cost is chosen to be the next \textit{current node} and the procedure is repeated until the horizon is reached, the whole graph is explored or the computation time limit for planning is reached. Finally, the node closest to the horizon is used to reconstruct the trajectory. %The pseudocode for this procedure is presented in Algorithm \ref{alg1}.
To avoid rounding errors, as the expansion of a node creates multiple transitions which in general do not end at gridpoints, the hybrid $A^{\ast}$ approach \cite{montemerlo2008junior_short} is used for planning. Hybrid $A^{\ast}$ also uses the grid, but keeps continuous values for the next expansion without rounding it to the grid, thus preventing the accumulation of rounding errors. 
Each \textit{node} $n$ contains 20 values: 6 indexes for each state in $\textbf{x}$ ($n.x_k$, $n.y_k$, $n.\psi_k$, etc.), 6 indexes for the \textit{parent node} (to reconstruct the trajectory), six remainders from the discretization of states ($n.x_r$, $n.y_r$, $n.\psi_r$, etc.), the exact \textit{cost-to-come} to the node ($n.g$), and the estimated total cost of traveling from the initial node to the goal region ($n.f$). The value $n.f$ is computed as $n.g + h(n)$, where $h(n)$ is the heuristic function.

The planning clearly requires processing time. The compensation of the planning time can be achieved by introducing $T_{plan}$, a guaranteed upper bound on planning time. The planning is then initiated from a position where the vehicle would be after the $T_{plan}$. The old trajectory is executed while the new one is being processed. Thus, the new trajectory is already planned when $T_{plan}$ arrives. This approach has been widely used in MP for automated vehicles \cite{ziegler2014trajectory}.

\subsection{Node expansion}
\label{subsec:node_exp}

To build trajectories iteratively, \textit{nodes} are expanded and \textit{child nodes} (final states on the motion primitives)  are generated, progressing towards the goal. From each node $n$, only dynamically feasible and collision-free \textit{child nodes} $n'$ should be generated.  As mentioned before, child nodes (motion primitives) are generated based on two models, bicycle model for a close-to-straight driving and vehicle-equilibrium-states during cornering.

%======================================%
\begin{figure}[!tp] 
\centering
\subfigure[]{%
\includegraphics[width=.43\columnwidth]{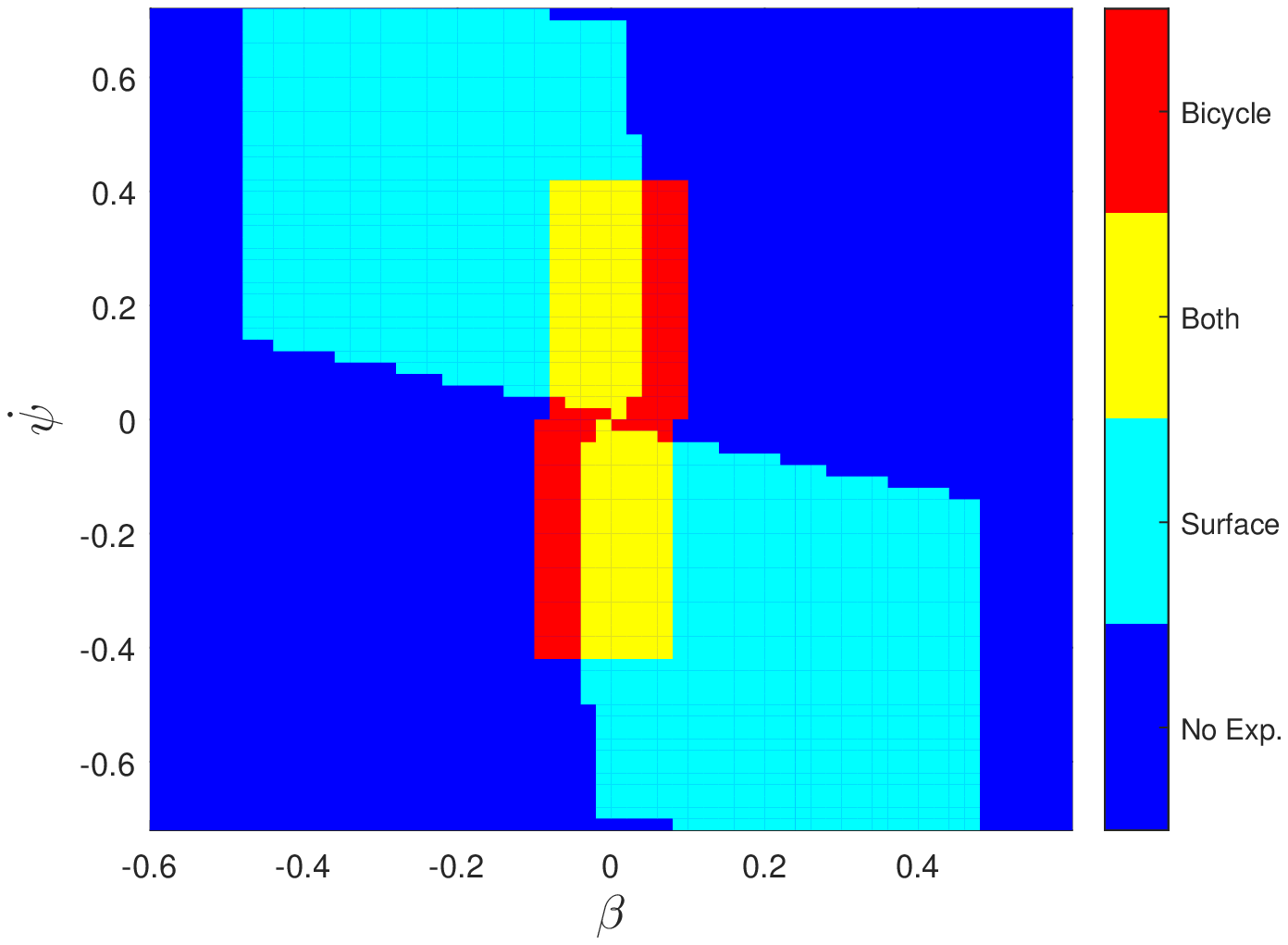}
\label{fig:expand1}%
}\hfil
\subfigure[]{%
\includegraphics[trim={1cm 0 1cm 0},clip,width=.55\columnwidth]{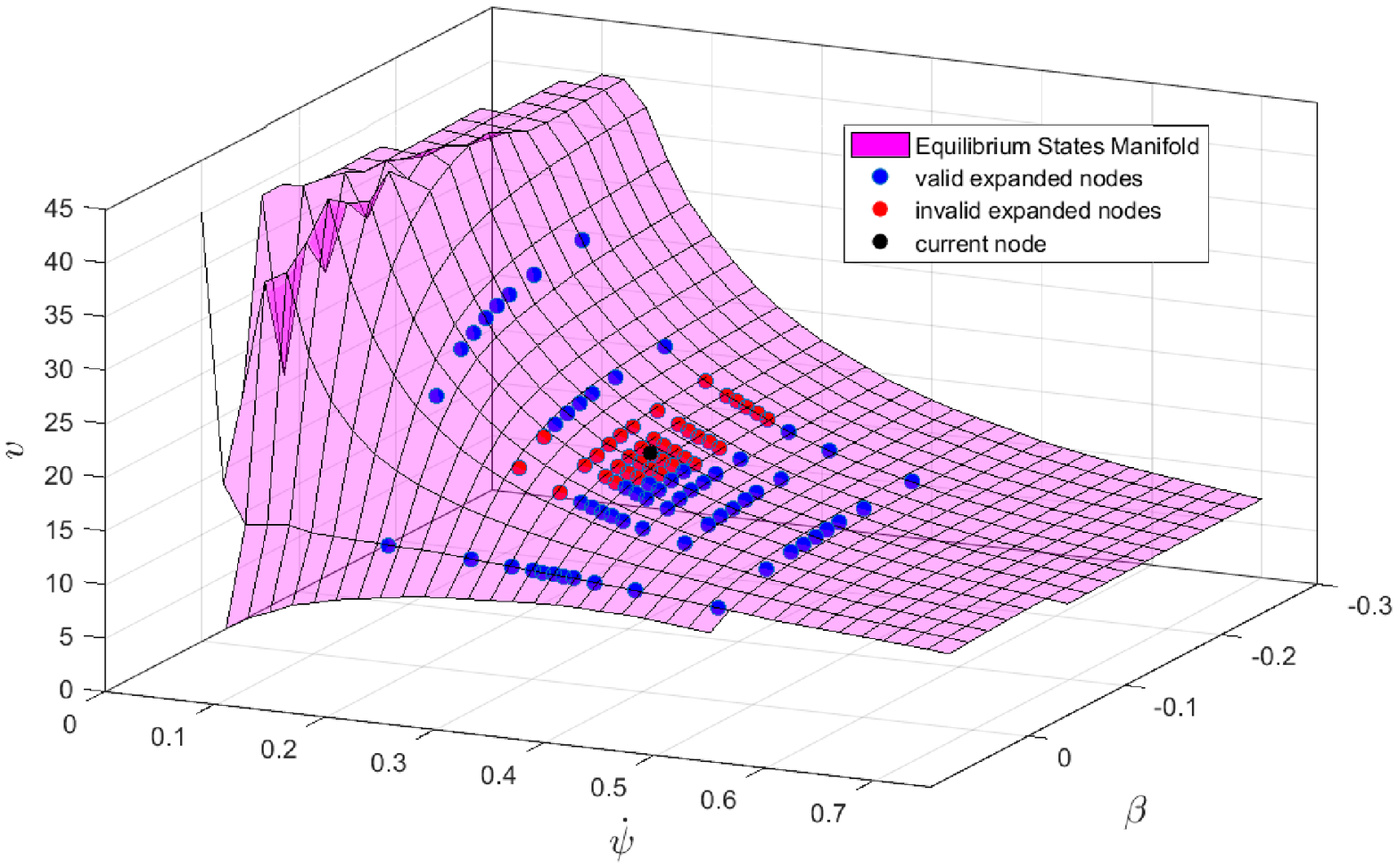}%
\label{fig:expand2}%
}
\caption{(a) \small $\beta-\dot\psi$ map representing the expansion modes depending on initial states. (b) \small Expanding parent node $n$ to different child nodes $n'$ on the ESM. }
\end{figure}
%======================================%

During cornering, the child nodes (motion primitives) are generated based on the steady state dynamic states $S_{ss}$, %( $v,\beta,\dot\psi$) 
as explained in Section \ref{subsec:v_manifold}. Based on the current steady state $S_{ss}$,  several reachable dynamic states are generated and \eqref{eq:kinematic_model} is used to simulate the evolution of other states ($x, y, \psi$) during the linear transition to these new steady states. 
%Still it is important to keep it low so limits on relative change must be set. % such as shown in (\ref{eq:Delta_assumption}). The smaller the deviations are, the closer the trajectory is to the ESM.
%Child nodes are created by considering linear transitions to several neighboring , while \eqref{eq:kinematic_model} is used to simulate the evolution of the states during this transition.
New steady states are obtained by sampling the $\beta-\dot\psi$ space around the current value $(\beta_0, \dot\psi_0)$, with the density of the samples decreasing as the distance from the current position increases (see Fig. \ref{fig:expand2}). The number of evaluated samples is a tuning parameter, which impacts considerably the performance of the search, as a trade-off between computation time and sufficient space exploration is needed. 
Since \eqref{eq:derivatives_0} assumes that the rate of change of states is equal to zero, in order to generate trajectories which keep the vehicle on the road, this constraint must be 'softened'.

In order to avoid generating and propagating an excessive amount of branches, before the nodes which are in collision are removed, the following rules are considered:
\begin{inparaenum}[i)]
	\item only equilibrium points defined within the surface in Fig. \ref{fig:surface_interp} are considered, so that the minimum reachable curvature radius is $R_{c,min}=10m$;
	\item the (small) portion of curve such that $\beta\cdot\dot{\psi}>0$ is neglected, since in general equilibrium points in which $\beta$ and $\dot\psi$ share the same sign are associated with low velocity conditions;
	\item a maximum velocity deviation $\Delta v$ between two successive nodes is defined, such that  $\Delta v/Ts<a_{max}$, where $a_{max}$ is the estimated maximum deceleration allowed on the given road surface.
\end{inparaenum}

Nodes are always expanded by \lq\lq exploring'' the neighboring states of the ESM. However, as illustrated in Fig. \ref{fig:expand1}, when the current conditions are close enough to the origin  of the $\beta-\dot\psi$ plane, i.e. for $|\beta|<\beta_{lin}$ and $|\dot\psi|<\dot\psi_{lin}$, also dynamic states and trajectories generated according to a nonlinear bicycle model are considered, where the forces in \eqref{eq:nonlinear_model} are replaced with their linearized approximations
\begin{equation}
\label{eq:bicycle_model_forces}
 F_{y,f} = -C_f(\beta+l_f\frac{\dot\psi}{v}-\delta) \mathrm{,}\quad F_{y,r}= -C_r(\beta-l_r\frac{\dot\psi}{v}) \mathrm{,}\quad F_{x,r} = -C_x \lambda
\end{equation}
%\begin{subequations}
%\label{eq:bicycle_model_forces}
%\begin{align}
 %F_{y,f} &= -C_f(\beta+l_f\frac{\dot\psi}{v}-\delta) \\
%F_{y,r} &= -C_r(\beta-l_r\frac{\dot\psi}{v}) \\
%F_{x,r} &= -C_x \lambda\end{align}
%\end{subequations}
In \eqref{eq:bicycle_model_forces} the longitudinal and lateral stiffness coefficients $C_x$, $C_f$, $C_r$ are consistent with the full characteristics given by \eqref{eq:mu_def}. In order to generate the node expansions, the steering wheel angle $\delta$ and the rear wheels slip $\lambda$ are varied within the ranges defined by $|\beta|<\beta_{lin}$ and $|\dot\psi|<\dot\psi_{lin}$ in the equilibrium surfaces in Fig. \ref{fig:surfaces_inputs}.

\subsection{Heuristic function}
The heuristic function $h(n)$ is used to estimate the cost needed to travel from some node $n$ to the goal state (\textit{cost-to-go}). As it is shown in \cite{hart1968formal}, if the heuristic function is underestimating the exact cost to go, the $A^{\ast}$ search provides the optimal trajectory. For the shortest path search, the usual heuristic function is the Euclidean distance. On the other hand, to find the minimum lap time, the heuristic should estimate the distance which the vehicle can travel from the current node during the defined time horizon. It is optimistic to assume that the vehicle accelerates (with maximum acceleration) in the direction of the road central line until it reaches the maximum velocity, and then maintains it for the rest of time horizon. Based on this velocity trajectory, the maximum travel distance can be computed and used as heuristic.

%Besides the expansion rules mentioned in \ref{subsec:node_exp}, 
Sacrificing optimality, in order to bias expansions towards the preferred motion and a better robustness, the heuristic function is augmented considering, among others:
\begin{inparaenum}[i)]
	\item a \lq\lq dynamic states evolution'' cost, which helps limiting the rate of change of the references $v$, $\beta$, $\dot\psi$, in order to obtain smooth trajectories and facilitate state tracking;
	\item penalization for trajectories approaching the road side;
	\item penalization of the nodes with less siblings, thus biasing the search to avoid regions where only few trajectories are feasible.
\end{inparaenum}

%%%%%%%%%% (IV) Simulation
\section{Simulations}

\label{sec:simulations}
%======================================%
\begin{figure}[!tp] 
\centering
\subfigure{
%\subfigure[U-turn]{%
\frame{\includegraphics[trim={190pt 85pt 136pt 54pt},clip, width=.45\columnwidth]{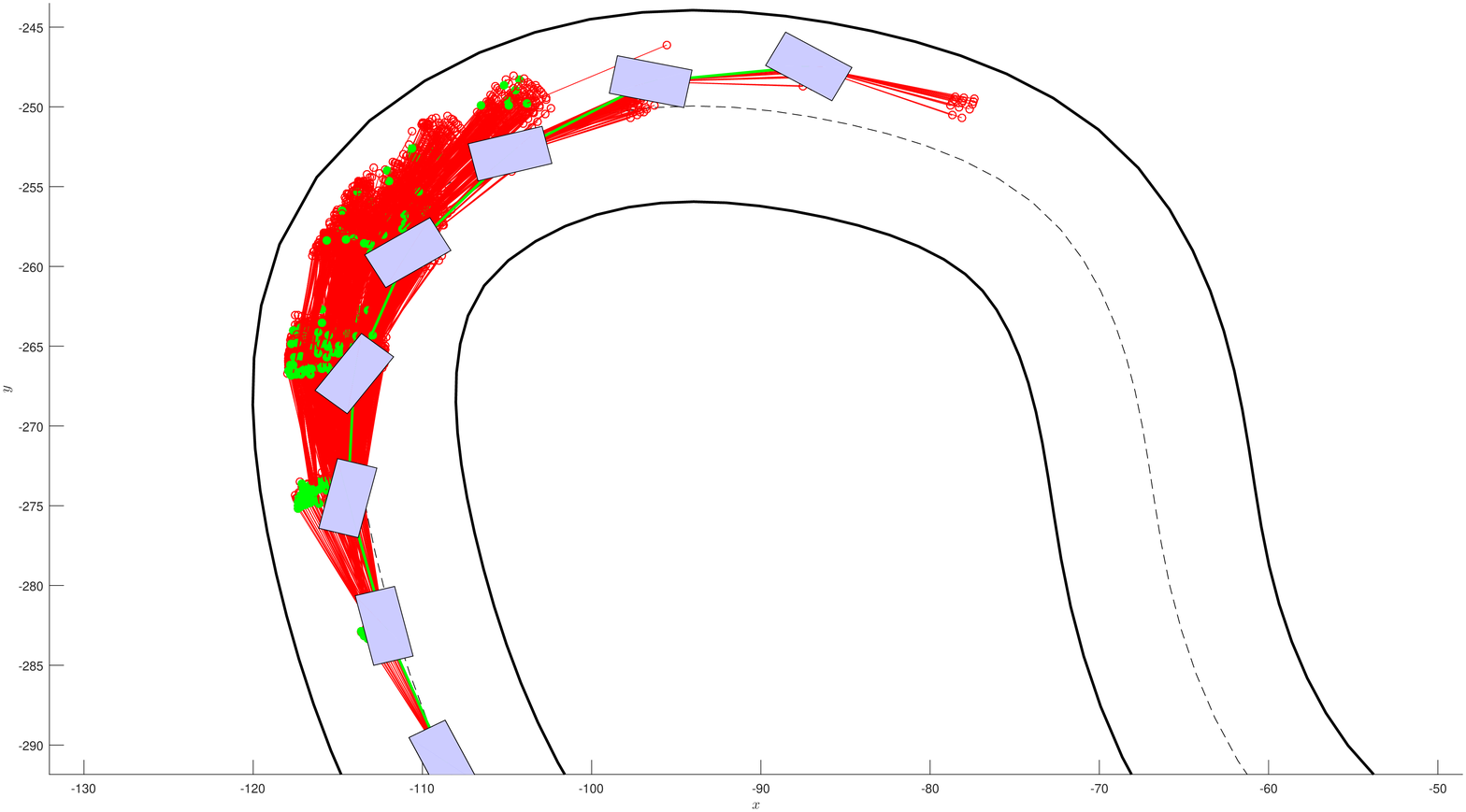}}%
%\label{fig:traj_exploration1}%
}\hfil
\subfigure{
%\subfigure[wide turn]{%
\frame{\includegraphics[trim={190pt 85pt 136pt 54pt},clip,width=.45\columnwidth]{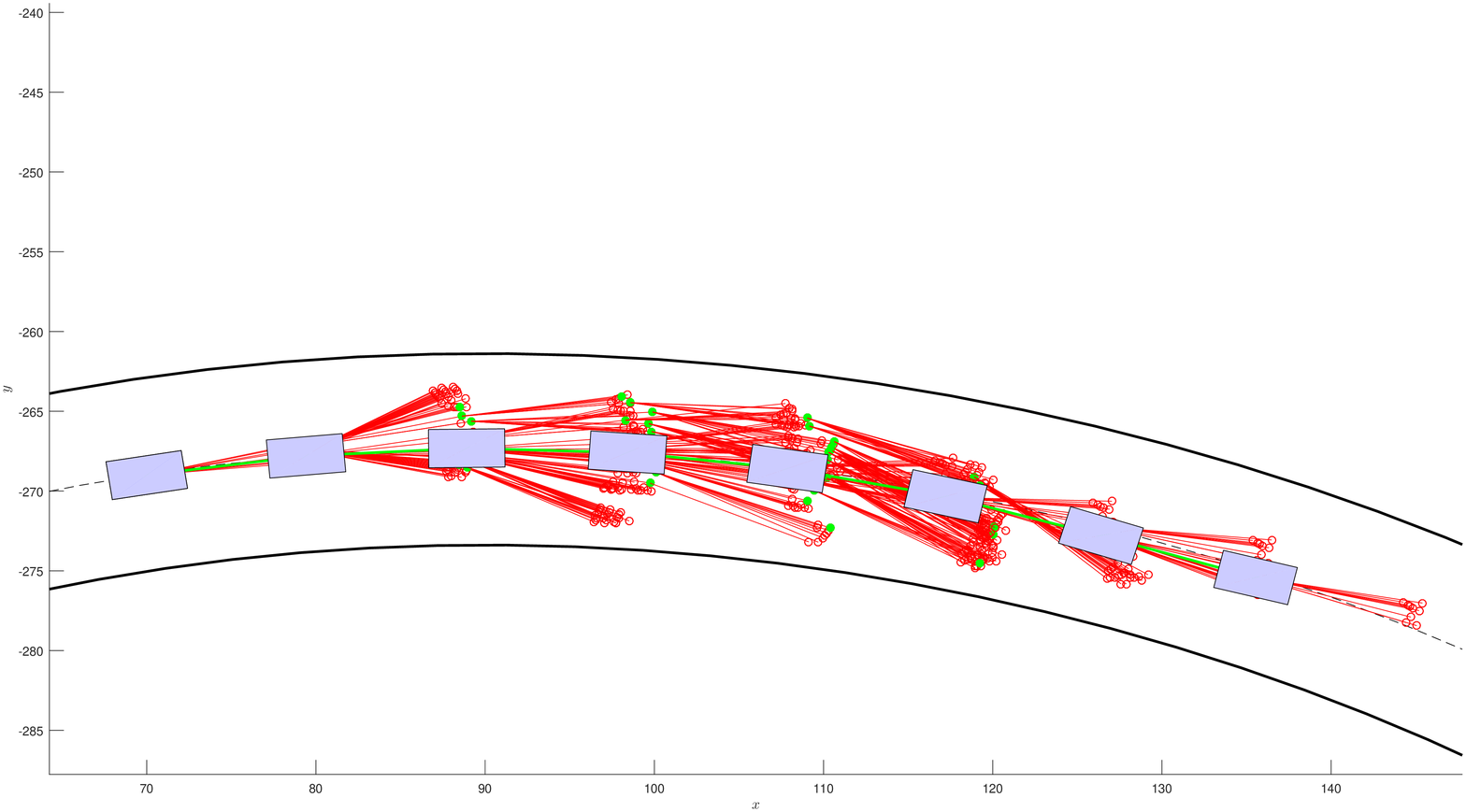}}
%\label{fig:traj_exploration2}%
}
\caption{Graphical representation of the trajectories exploration in U-turn (left) and wide turn (right).}
\label{fig:traj_exploration}%
\end{figure}
%======================================%

The presented concept is evaluated in the Matlab/SIMULINK environment, assuming perfect actuation, i.e. the actual vehicle dynamical states/positions match the ones planned at the previous iteration.
%$x = x^{\ast}$, for $x = v, \beta, \dot\psi$.
The trajectory exploration can be appreciated in Fig. \ref{fig:traj_exploration} in the case of a U-turn and of a wider curve.
The explored branches are represented by the red links, and the closed nodes are marked as green. The light-blue car frames represent the optimal vehicle states (see Fig. \ref{fig:traj_exploration}).

\begin{figure}[!tp] 
\centering
\subfigure[U-turn maneuver: consecutive frames.]{
\begin{tabular}{|c|c|c|c|c|c|c|}
      \hline
      \includegraphics[trim={11cm 0 11cm 0},clip,width=.125\columnwidth]{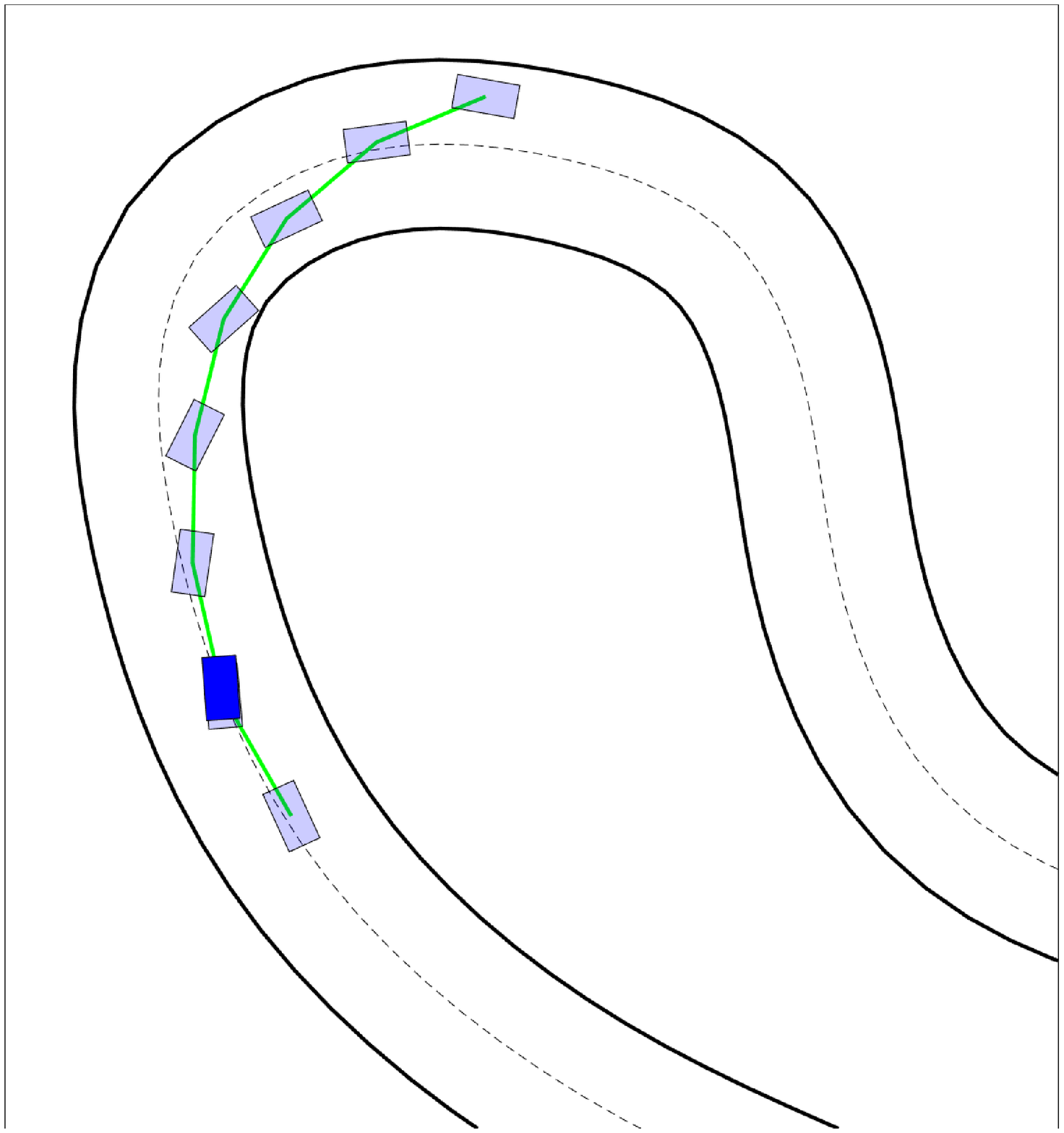} &
	  \includegraphics[trim={11cm 0 11cm 0},clip,width=.125\columnwidth]{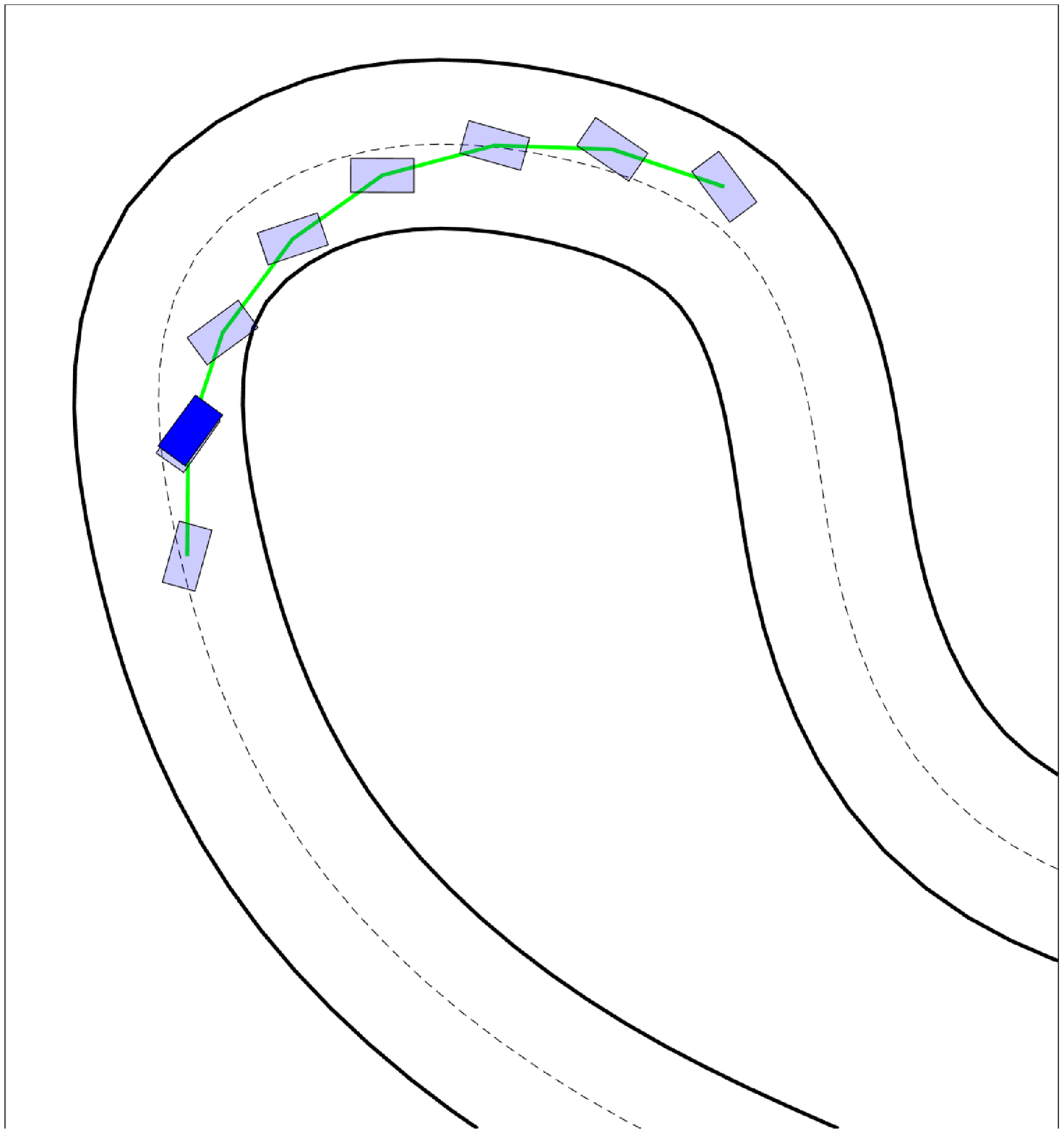} &
      \includegraphics[trim={11cm 0 11cm 0},clip,width=.125\columnwidth]{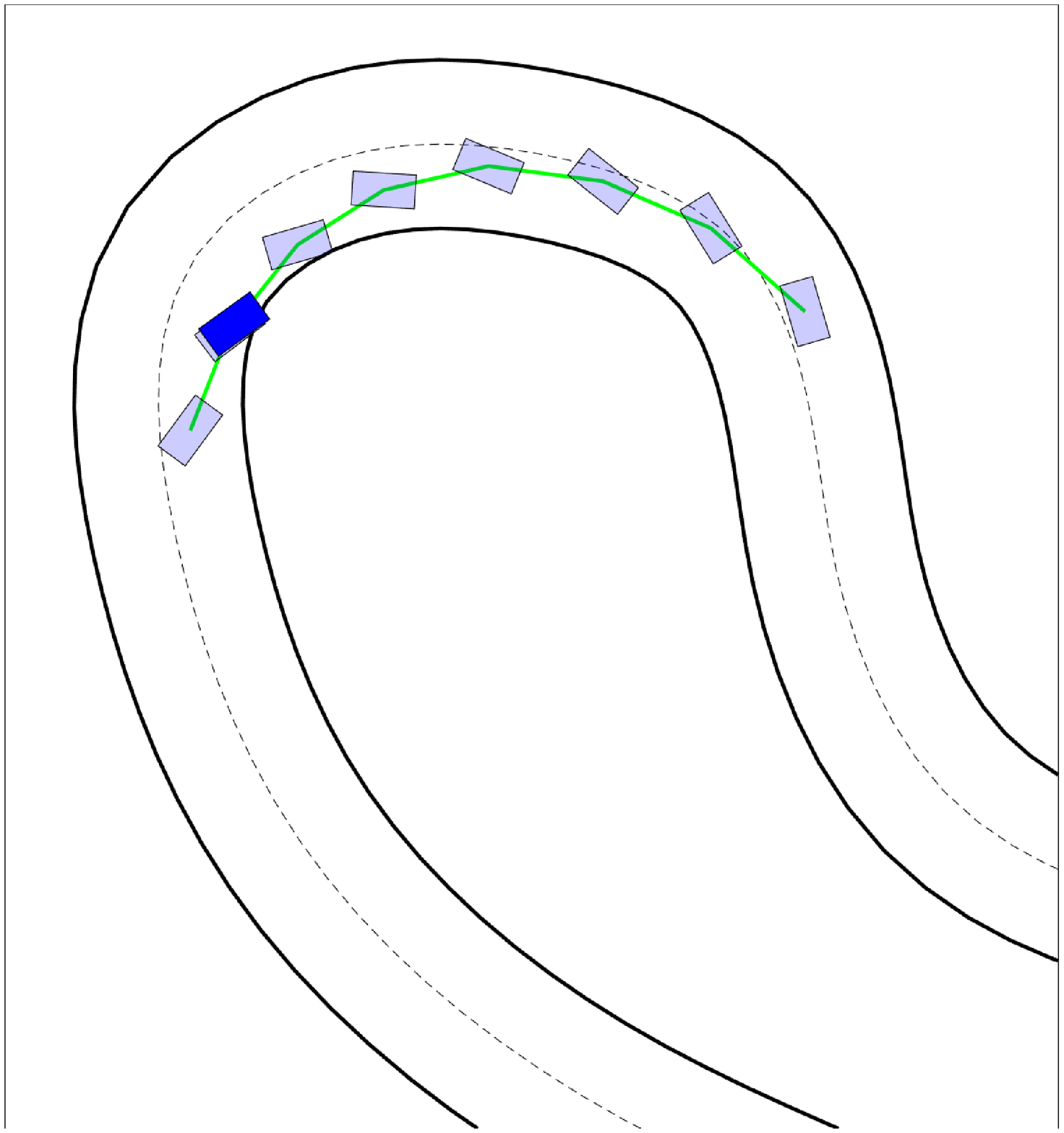} &
      \includegraphics[trim={11cm 0 11cm 0},clip,width=.125\columnwidth]{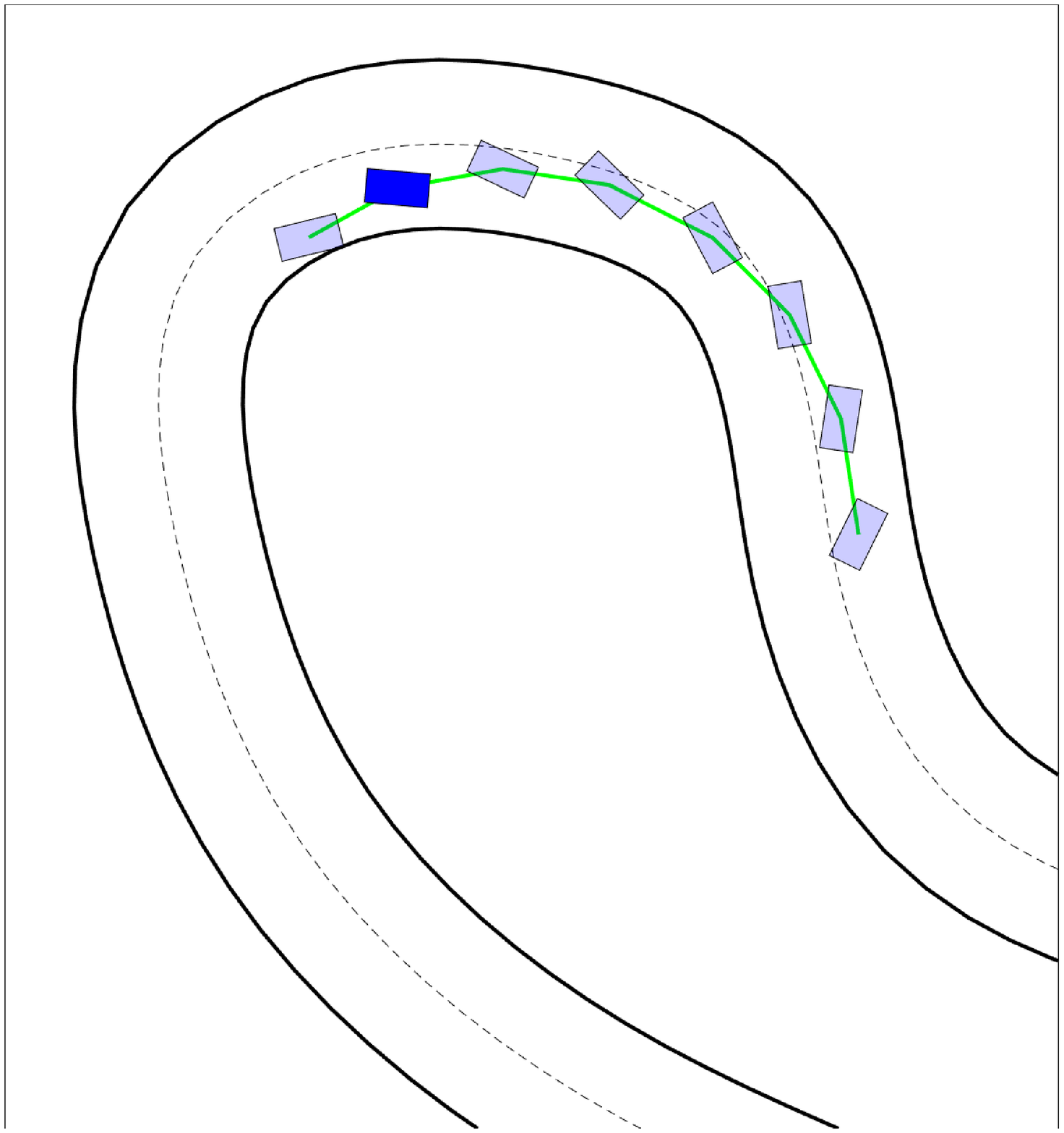} &
      \includegraphics[trim={11cm 0 11cm 0},clip,width=.125\columnwidth]{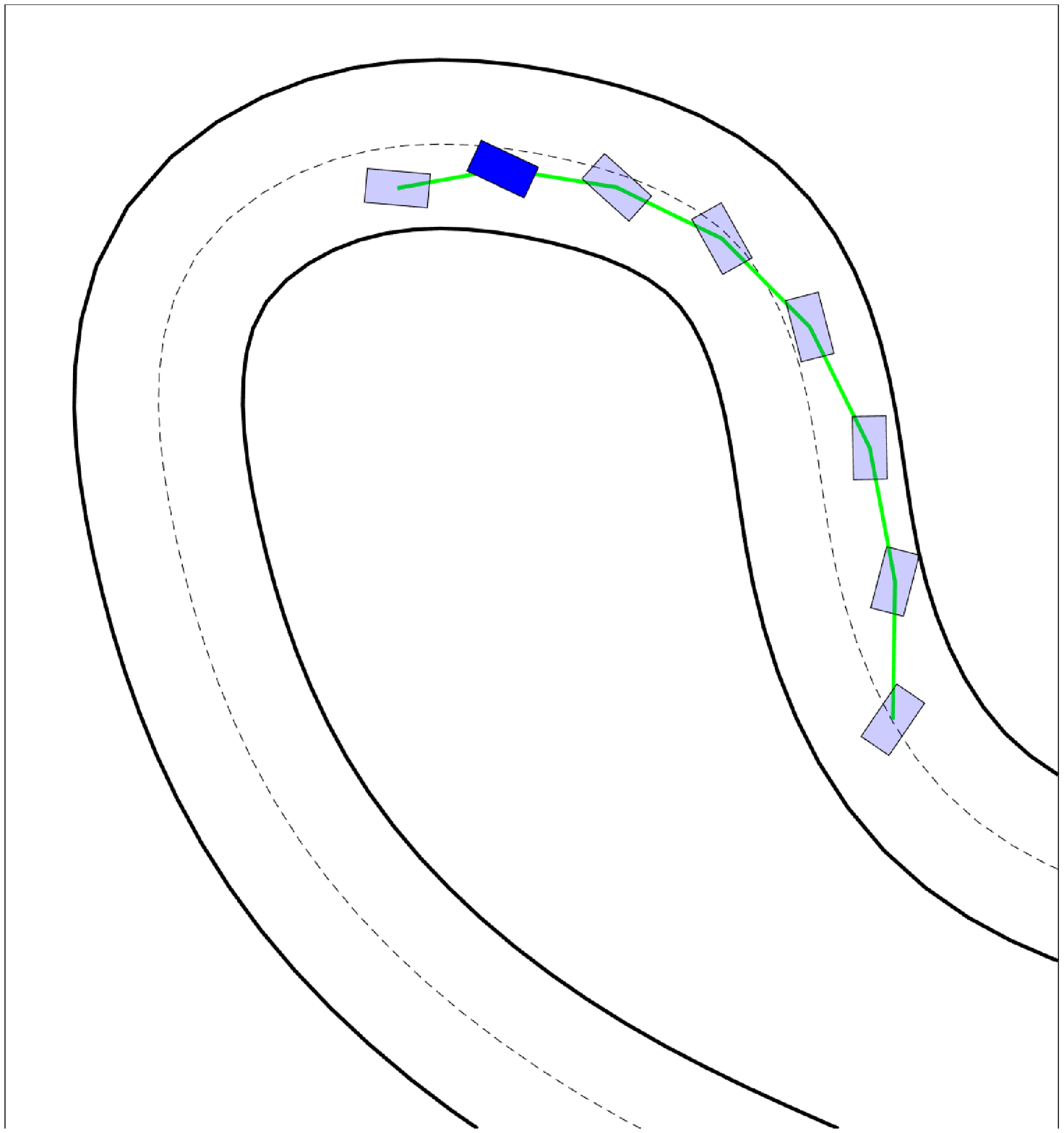} &
	  \includegraphics[trim={11cm 0 11cm 0},clip,width=.125\columnwidth]{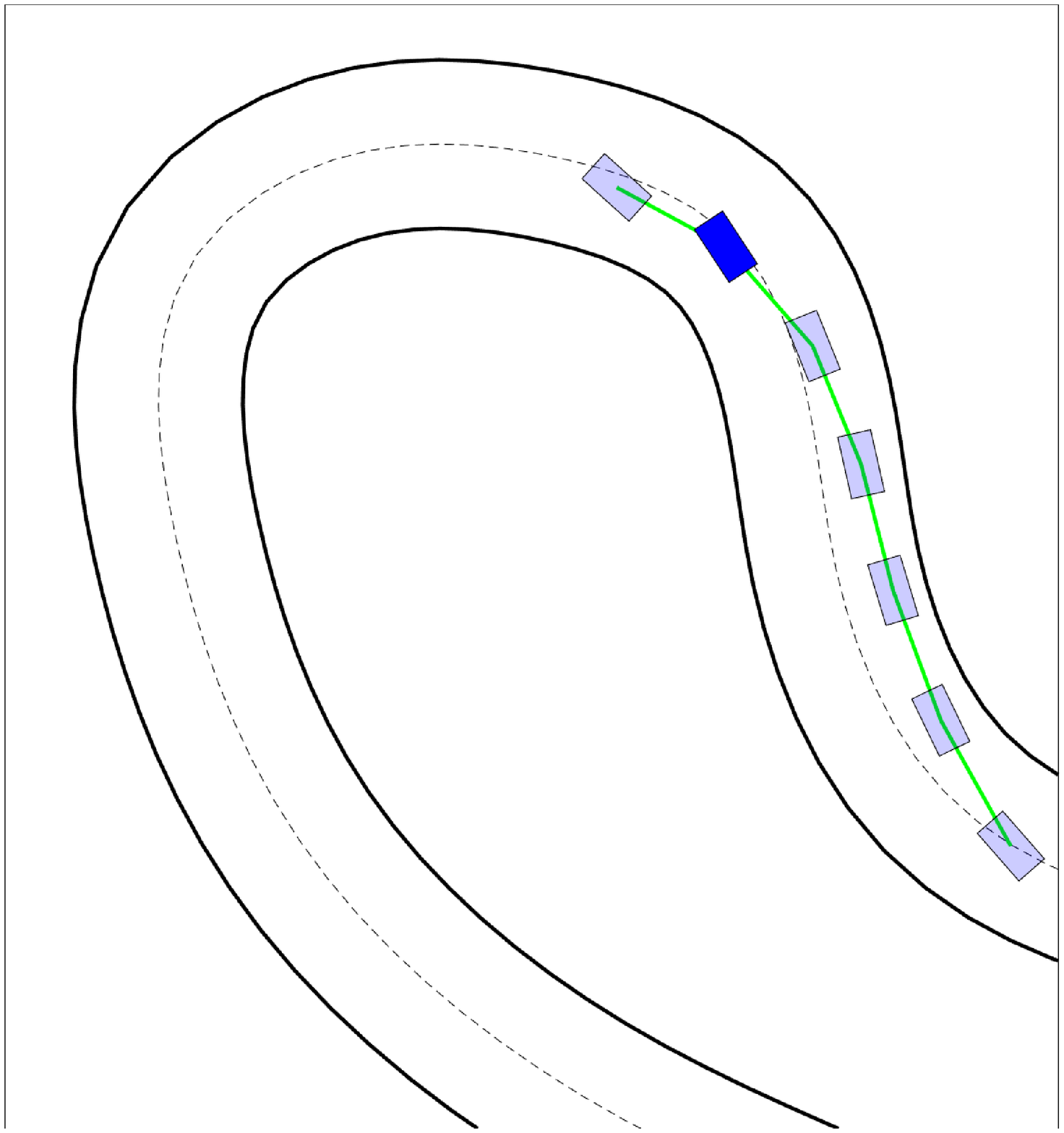} &
	  \includegraphics[trim={11cm 0 11cm 0},clip,width=.125\columnwidth]{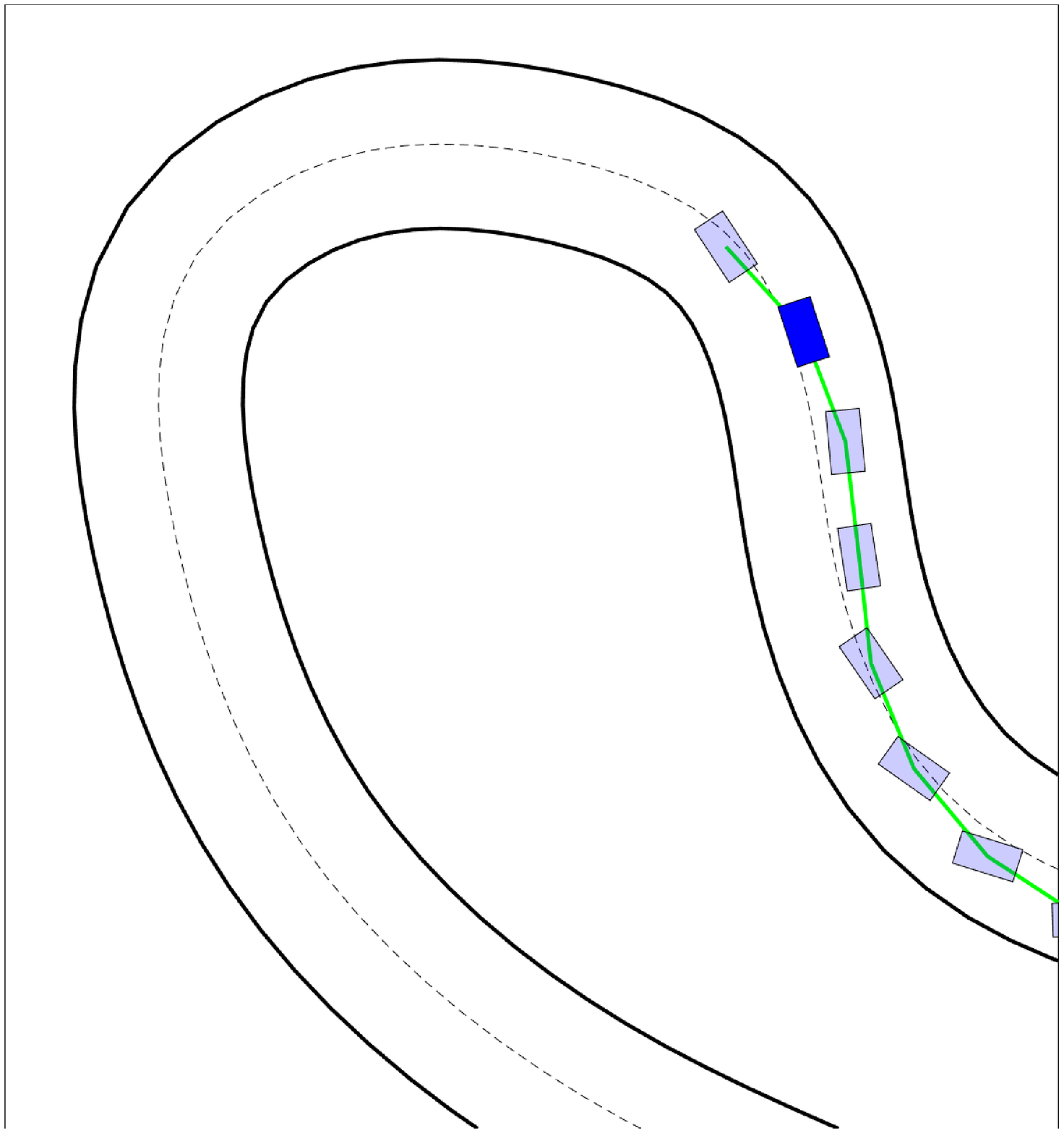} \\						
      \hline
      \multicolumn{7}{c}{}
\end{tabular}
\label{fig:frames}%
}
\subfigure[Reference dynamical states.]{%
\includegraphics[trim={0.7cm 0.2cm 1.5cm 0.3cm},clip,width=.42\columnwidth]{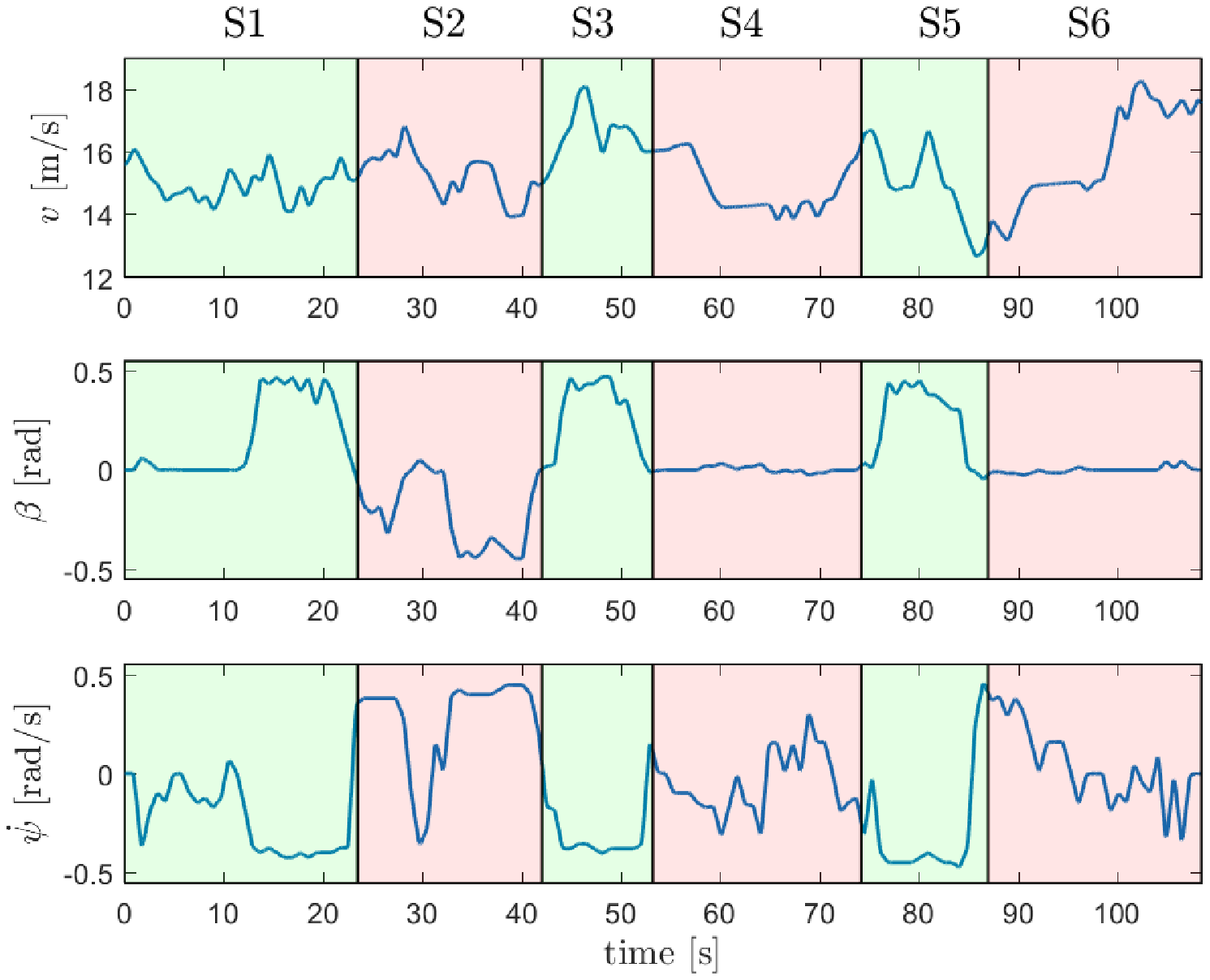}%
\label{fig:states_graph}%
}\hfil
\subfigure[Obtained driving trajectory.]{%
\includegraphics[trim={2.5cm 0.5cm 2.5cm -0.5cm},clip,width=.52\columnwidth]{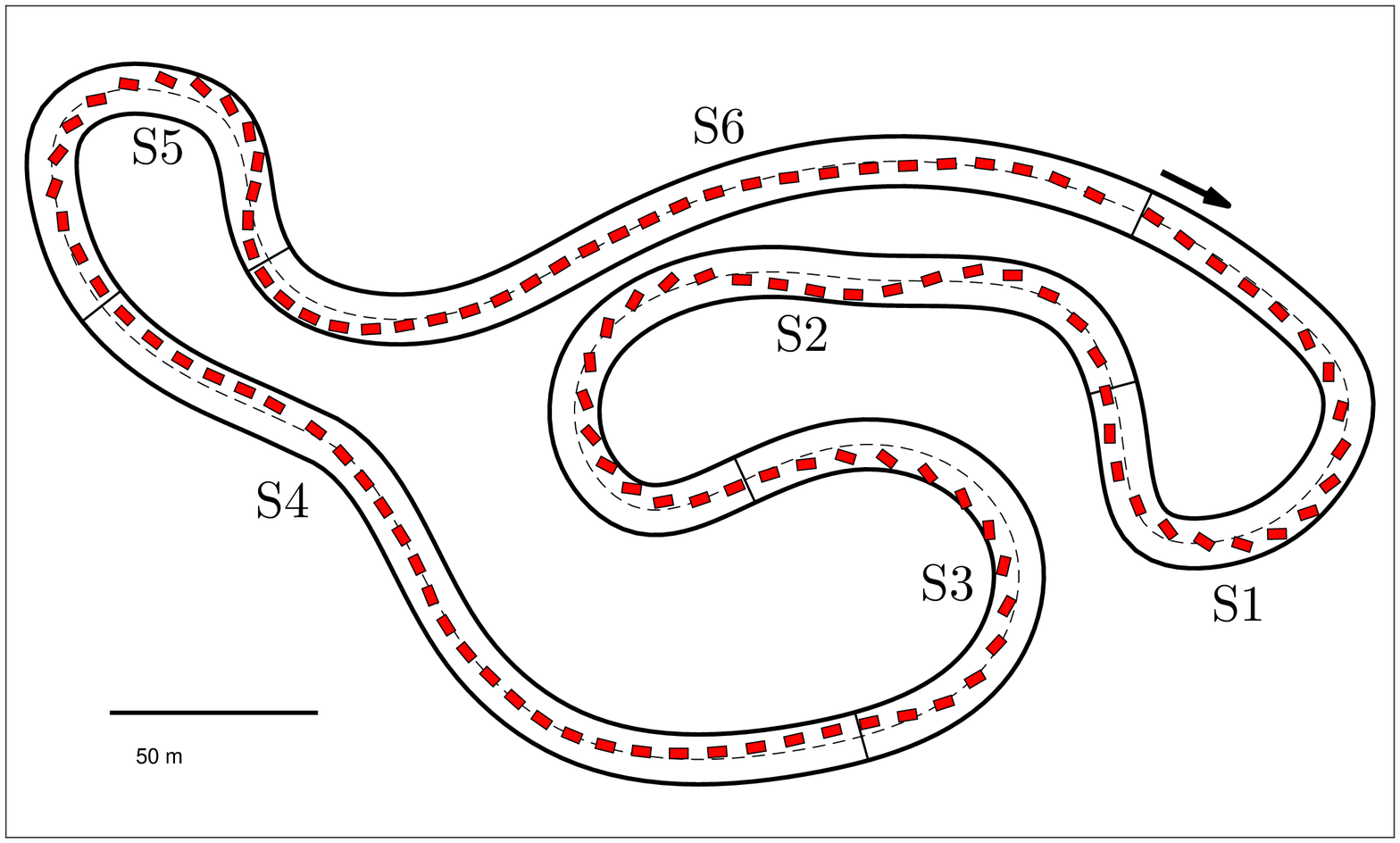}
\label{fig:trajectories}%
}
\caption{Dynamical states  and trajectory evolution over the full test-circuit.}
\end{figure}

In Fig. \ref{fig:frames} several frames of the same maneuver are shown (the top left turn in the track illustrated in Fig. \ref{fig:trajectories}). From these, it is possible to appreciate how at each iteration the optimal trajectory is re-calculated based on the current position. Given the nature of the receding horizon approach, it is not guaranteed (nor preferred) that all or part of the previously computed trajectory are kept in the next iteration. In fact, while in the first step the trajectory 'dangerously' approaches the side of the road, in the next two steps the trajectory is incrementally regularized, thanks to the fact that the exploration of such portion of the track is now being evaluated in earlier nodes.

The dynamical states, which represent the output of the trajectory generation, are depicted in Fig. \ref{fig:states_graph}. One can see how the generated references are varied smoothly, in particular in terms of $v$ and $\beta$, which are the quantities characterized by slower actuation dynamics.
Moreover, it is possible to distinguish clearly 4 intervals in which the optimal generated maneuver is a 'drift' one with $\beta>0.4 rad$.
These same intervals can be recognized in Fig. \ref{fig:trajectories}, where the overall trajectory on the considered $10m$-wide track can be evaluated.

%======================================%
%======================================%

%%%
%%%%%%%%%% (V) Results
\section{Conclusions}
\label{sec:results}
%\lipsum[9]

In this paper, a novel $A^{\ast}$ search-based motion planning for performance driving is presented and used to generate dynamically feasible trajectories on a slippery surface. The proposed method extends drift-like driving from a steady state drifting on single curve to a continuous driving on the road effectively entering and exiting drifting maneuvers and switching between right and left turns.
The proposed method assumes that the vehicle parameters and the road surface properties are known to a certain degree, which allows to define a set of steady-state cornering maneuvers. %, defined as ESM.
The method is evaluated on a mixed circuit characterized by slippery conditions (gravel), which contains several road sections of varying curvature radii $R_c$. 
In several instances, due to the particular road surface considered, the optimal selected trajectory involves drifting, which in certain conditions ensures the maximum lateral acceleration.
Such a result demonstrates the capability of the proposed hybrid $A^{\ast}$ algorithm to generate feasible sub-optimal trajectories on slippery conditions, while considering a limited prediction horizon. Moreover, when considering U-turns with curvature radius as tight as 15m, such trajectories are comparable in shape to the ones obtained when the full segment is optimized in order to find the minimum time optimal maneuver, as in e.g. \cite{tavernini2013minimum}.
%
%

%% ACKNOWLEDGEMENTS
%%%%%%%%%%%%%%%%%%%%%%%%%%%%%%%%%%%%%%%%%%%%%%%%%%%%%%%%%%%%%%%%%%%%%%%%%%%%%%%%
\section*{Acknowledgment}
\begin{footnotesize}
The project leading to this study has received funding from the European Union's Horizon 2020 research and innovation programme under the Marie Sk\l{}odowska-Curie grant agreement No 675999, ITEAM project.
VIRTUAL VEHICLE Research Center is funded within the COMET - Competence Centers for Excellent Technologies - programme by the Austrian Federal Ministry for Transport, Innovation and Technology (BMVIT), the Federal Ministry of Science, Research and Economy (BMWFW), the Austrian Research Promotion Agency (FFG), the province of Styria and the Styrian Business Promotion Agency (SFG). The COMET programme is administrated by FFG.
\end{footnotesize}
%%%%%%%%%%%%%%%%%%%%%%%%%%%%%%%%%%%%%%%%%%%%%%%%%%%%%%%%%%%%%%%%%%%%%%%%%%%%%%%%

\bibliographystyle{IEEEtran}
\bibliography{biblio}
%\balance

\end{document}